%% file: icml-main.tex
\documentclass{article}

\usepackage{microtype}
\usepackage{graphicx}
\usepackage{subfigure}
\usepackage{booktabs} 

\usepackage{hyperref}

\usepackage[accepted]{icml2024}

\usepackage{amsmath}
\usepackage{amssymb}
\usepackage{mathtools}
\usepackage{amsthm}
\usepackage{stmaryrd}
\usepackage{turnstile}
\usepackage{bm} 

\usepackage[capitalize,noabbrev]{cleveref}

\theoremstyle{plain}
\newtheorem{theorem}{Theorem}[section]

\newtheorem{lemma}[theorem]{Lemma}
\newtheorem{corollary}[theorem]{Corollary}
\theoremstyle{definition}
\newtheorem{definition}[theorem]{Definition}

\theoremstyle{remark}
\newtheorem{remark}[theorem]{Remark}

\newtheorem*{theorem*}{Theorem}
\newtheorem*{lemma*}{Lemma}
\newtheorem{fact}[theorem]{Fact}
\newtheorem*{fact*}{Fact}
\newtheorem*{proposition*}{Proposition}
\newtheorem*{corollary*}{Corollary}

\newtheorem*{hypothesis*}{Hypothesis}

\newtheorem*{conjecture*}{Conjecture}
\theoremstyle{definition}
\newtheorem*{definition*}{Definition}

\newtheorem*{construction*}{Construction}

\newtheorem*{example*}{Example}

\newtheorem*{question*}{Question}
\newtheorem*{assumption*}{Assumption}

\newtheorem*{problem*}{Problem}

\newtheorem*{model*}{Model}
\theoremstyle{remark}

\newtheorem*{claim*}{Claim}
\newtheorem*{remark*}{Remark}

\newtheorem*{observation*}{Observation}

\usepackage[textsize=tiny]{todonotes}

\newcommand{\Authornote}[2]{{\sffamily\small\color{red}{[#1: #2]}}}

\newcommand{\Tnote}{\Authornote{Tommaso}}

\usepackage{boxedminipage}

\newcommand{\Paren}[1]{\left(#1\right)}

\newcommand{\brac}[1]{[#1]}
\newcommand{\Brac}[1]{\left[#1\right]}

\newcommand{\card}[1]{\lvert#1\rvert}

\newcommand{\set}[1]{\{#1\}}
\newcommand{\Set}[1]{\left\{#1\right\}}

\newcommand{\norm}[1]{\lVert#1\rVert}
\newcommand{\Norm}[1]{\left\lVert#1\right\rVert}

\newcommand{\Normt}[1]{\Norm{#1}_2}

\newcommand{\Snormt}[1]{\Norm{#1}^2_2}

\newcommand{\Snorm}[1]{\Norm{#1}^2}

\newcommand{\Normo}[1]{\Norm{#1}_1}

\newcommand{\Normi}[1]{\Norm{#1}_\infty}

\newcommand{\Normn}[1]{\Norm{#1}_\textnormal{nuc}}

\newcommand{\Normm}[1]{\Norm{#1}_\textnormal{max}}

\newcommand{\iprod}[1]{\langle#1\rangle}

\newcommand{\Esymb}{\mathbb{E}}

\DeclareMathOperator*{\E}{\Esymb}

\newcommand{\suchthat}{\;\middle\vert\;}
\newcommand{\tensor}{\otimes}

\newcommand{\sge}{\succeq}

\newcommand{\tensorpower}[2]{#1^{\tensor #2}}

\newcommand{\from}{\colon}

\newcommand{\mper}{\,.}

\newcommand\bdot\bullet

\DeclareMathOperator{\Tr}{Tr}

\newcommand{\N}{\mathbb N}
\newcommand{\R}{\mathbb R}

\newcommand{\cA}{\mathcal A}
\newcommand{\cB}{\mathcal B}

\newcommand{\cD}{\mathcal D}
\newcommand{\cE}{\mathcal E}

\newcommand{\cG}{\mathcal G}

\newcommand{\cM}{\mathcal M}

\newcommand{\cO}{\mathcal O}
\newcommand{\cP}{\mathcal P}
\newcommand{\cQ}{\mathcal Q}

\newcommand{\cS}{\mathcal S}

\newcommand{\cV}{\mathcal V}

\newcommand{\cX}{\mathcal X}
\newcommand{\cY}{\mathcal Y}

\newcommand{\bbP}{\mathbb P}

\renewcommand{\leq}{\leqslant}
\renewcommand{\le}{\leqslant}
\renewcommand{\geq}{\geqslant}
\renewcommand{\ge}{\geqslant}
\let\epsilon=\varepsilon

\newcommand{\eps}{\epsilon}

\allowdisplaybreaks
\newcommand*{\Id}{\mathrm{Id}}

\newcommand*{\Normf}[1]{\Norm{#1}_{\mathrm{F}}}

\newcommand*{\transpose}[1]{{#1}{}^{\mkern-1.5mu\mathsf{T}}}
\newcommand*{\dyad}[1]{#1#1{}^{\mkern-1.5mu\mathsf{T}}}

\renewcommand{\ij}{{ij}}

\icmltitlerunning{Perturb-and-Project: Differentially Private Similarities and Marginals}

\begin{document}
\onecolumn

\icmltitle{Perturb-and-Project: Differentially Private Similarities and Marginals}



\icmlsetsymbol{equal}{*}

\begin{icmlauthorlist}
\icmlauthor{Vincent Cohen-Addad}{equal,google}
\icmlauthor{Tommaso d'Orsi}{equal,google,bocconi}
\icmlauthor{Alessandro Epasto}{equal,google}
\icmlauthor{Vahab Mirrokni}{equal,google}
\icmlauthor{Peilin Zhong}{equal,google}
\end{icmlauthorlist}

\icmlaffiliation{google}{Google Research}
\icmlaffiliation{bocconi}{BIDSA, Bocconi}

\icmlcorrespondingauthor{Tommaso d'Orsi}{tommaso.dorsi@unibocconi.it}
\icmlcorrespondingauthor{Alessandro Epasto}{aepasto@google.com}

\icmlkeywords{Differential privacy, Machine Learning, ICML, k-way marginals, pair-wise similarities, sum-of-squares}

\vskip 0.3in



\printAffiliationsAndNotice{\icmlEqualContribution} 

\begin{abstract}

We revisit the input perturbations framework for differential privacy where noise is added to the input $A\in \cS$ and the result is then projected back to the space of admissible datasets $\cS$. 
Through this framework, we first design novel efficient algorithms to privately release pair-wise cosine similarities.
Second, we derive a novel algorithm to compute $k$-way marginal queries over $n$ features. Prior work could achieve comparable guarantees only for $k$ even. Furthermore, we extend our results to $t$-sparse datasets, where our efficient algorithms yields novel, stronger guarantees whenever $t\le n^{5/6}/\log n\,.$
Finally, we provide a theoretical perspective on why \textit{fast} input perturbation algorithms works well in practice.
The key technical ingredients behind our results are tight sum-of-squares certificates upper bounding the Gaussian complexity of sets of solutions.
\end{abstract}

\input{introduction}

\input{results}
\input{related}
\input{preliminaries}
\input{techniques}
\input{meta-theorem}

\input{distances-release}

\section*{Acknowledgments}
The authors thank A. Nikolov for useful discussions concerning $k$-way marginal queries and the perturb-and-project framework.
Tommaso d’Orsi is partially supported by the project MUR FARE2020 PAReCoDi.

\bibliography{custom}
\bibliographystyle{icml2024}

\newpage
\appendix
\onecolumn

\input{k-way-marginals}
\input{background}

\end{document}

%% file: introduction.tex
\section{Introduction}
Differential privacy (DP)~\cite{dwork2014algorithmic} has become the golden standard to define how much information an algorithm leaks about users' data. Thus, designing differentially private algorithms is a central
problem in modern machine learning. One of the key considerations in designing DP algorithms is to determine the minimum amount of noise needed to ensure privacy to maximize the accuracy of the output: Insufficient noise may result in a non-private algorithm while excessive noise can significantly degrade the algorithm's output quality. It is thus key to quantify how much a worst-case user impacts~\cite{dwork2006calibrating} the final output of the algorithm to add the appropriate noise.



An interesting example is the case of empirical risk minimization (ERM) problems. Iterative optimization algorithms such as stochastic gradient descent (SGD) are central algorithms in modern machine learning. However, making these algorithms differentially private is
particularly challenging~\cite{chaudhuri2011differentially}; repeated noise addition is necessary, leading to an accumulation of noise that can significantly degrade the final solution's quality. Furthermore, precisely quantifying the required noise at each iteration is often difficult, potentially resulting in overly conservative noise addition.

Another widely-used approach to ERM problems is objective perturbation~\cite{chaudhuri2008privacy,chaudhuri2011differentially,iyengar2019towards,kifer2012private}, which involves perturbing the 
objective function so that optimizing the perturbed objective ensures that the output is
private. 
One very generic way of achieving objective perturbation is through \emph{input 
perturbation} which consists in adding noise to the input dataset to obtain a private
perturbed input. This permits the use of any non-DP algorithms on the perturbed input and hence simplifies practical implementation. One benefit over perturbing the objective is that
the properties (e.g., convexity) of the objective are unchanged and so the state-of-the-art
non-DP optimizers can be used. 
Experimentation with various non-DP algorithms becomes possible, and privacy guarantees are immediate. A crucial research direction is thus to investigate input perturbation methods that preserve the intrinsic properties of the data that are beneficial for the downstream task.

\paragraph{Our Contribution}
In this work, we study input perturbation techniques from a theoretical perspective. 
We focus on the problem of designing differentially private projection functions (such as e.g., dimensionality reduction techniques). We observe that, quite surprisingly, input perturbations yield the best known (and conjecturally optimal) guarantees for a large class of projection functions.
This challenges the expectation that a general approach might incur sub-optimal utility loss. Our analysis reveals that the algorithm's utility guarantees are contingent upon the ``richness'' (metric entropy) of the target set (the set into which the input is projected). Inspired by~\cite{d2023higher}, we use sum-of-squares to obtain tight bounds on the Gaussian 
complexity of set of solutions. 
To further improve, we introduce new sum-of-square certificates 
(e.g. for sparse injective tensor norms). 
Finally, while our sum-of-squares solution might appear complex, we extract an interesting explanation for the practical success of algorithms like linear projection.  At a high-level, we exploit the fact that the sum-of-squares projections onto convex sets can be broken down into projections onto simpler convex sets (this is a consequence of results
on alternating projections, see \cite{bauschke1993convergence, sakai1995strong}).
The result is that for polynomially bounded inputs, this can thus be achieved by using  a small (often only logarithmic)
number of basic projections, a popular approach in practice.

%% file: results.tex
\section{Results}\label{section:results}
\paragraph{Privately releasing pair-wise distances}
Our first result concerns the task of privately releasing pair-wise cosine similiarities of a set of $n$ vectors~\cite{yang2017privacy,fernandes2021locality}.
This is a fundamental subroutine of many algorithms (e.g. nearest neighbors search).
Without any constraint on the vectors structure, one should not expect to output a reasonable estimate without sacrificing privacy. This is because even a single change in a single entry of a vector can drastically alter all its inner products (for example if the vector is sparse).
We consider the following natural notion of adjacency over matrices $V\,, V'\in \R^{n\times m}\,,$ related notions appeared in \cite{blocki2012johnson, imola2023differentially},
\begin{align*}
    \Delta &\geq
    \sum_\ij \Paren{\iprod{V_i,V_j}-\iprod{V'_i,V'_j}}^2\\ 
    &=\Normf{\dyad{V}-\dyad{V'}}\,.
\end{align*}
That is, the sum of the squared differences between the inner products of the rows of $V$ and $V'$ is bounded by some chosen parameter $\Delta\,.$ Notice that, for example,  for set of vectors in $\set{\pm 1/\sqrt{n}}^n\,,$ this definition captures $O(\sqrt{n})$ entry-wise changes.

\begin{theorem}\label{theorem:cosine-similarities-release}
    Let $V=:\Set{v_1,\ldots,v_n}\subseteq \R^m$ be a set of unit vectors. 
    There exists an $(\eps,\delta)$-differentially private algorithm that, on input $V$, returns a matrix $\hat{\mathbf X}\in \R^{n\times n}$ satisfying\footnote{We use boldface to denote random variables.}
    \begin{align*}
        \E   \Normf{\dyad{V}-\hat{\mathbf{X}}}^2
        \leq O\Paren{\frac{\Delta\cdot \sqrt{\log (2/\delta)}}{\eps}}\cdot n^{3/2}\,.
    \end{align*}
    Moreover, the algorithm runs in polynomial time.
\end{theorem}
Prior works focused either on vector level differential privacy for restricted classes of vectors  \cite{kenthapadi2012privacy, blocki2012johnson}, 
or considered local extended differential privacy \cite{fernandes2021locality}, making a fair comparison somewhat hard. 
In the latter case, when extended to standard differential privacy, the resulting guarantees are worse than \cref{theorem:cosine-similarities-release}.
In the former, we observe that \cite{kenthapadi2012privacy} improves over the naive Gaussian mechanism only for sparse vectors with $o(n)$ non-zero entries. In contrast,  compared to \cref{theorem:cosine-similarities-release} the Gaussian mechanism would yield guarantees worse by a $\sqrt{n}$ factor, for any set of vectors 
Moreover, as we discuss in \cref{remark:distance-releases-optimality}, evidence suggests the bound of \cref{theorem:cosine-similarities-release} may be information theoretically optimal.

\paragraph{K-way marginals}
We also apply our framework to privately compute $k$-way marginal queries~\cite{kasiviswanathan2010price}. Roughly, speaking, in these settings, the input is a dataset of vectors in $\set{0, 1}^n$ and the goal is to answer queries of the form: \textit{"How many vectors are non-zero in the entries in $V\subseteq [n]\,, \card{V}\leq k$?}
The definition of adjacency here is the natural notion in which one vector in the dataset is replaced.
For $k$-way marginal queries we obtain the following two results (formally stated in \cref{section:k-way-marginals}). 

\begin{theorem}[K-way marginals, informal]\label{theorem:k-way-marginals-informal}
Let $D$ be a dataset of $m$ elements, each a binary vector in $\set{0,1}^{n}\,.$ There exists an $(\epsilon,\delta)$-differentially private algorithm that,  on input $D\,,k$, answers all $k$-way marginal queries with expected average query-wise squared error 
\begin{align*}
    O\Paren{m\cdot n^{k/4}\cdot \frac{\sqrt{\log (2/\delta)}}{\eps}}
\end{align*}
when $k$ is even,
\begin{align*}
    O\Paren{m\cdot (n^{k}\cdot k \cdot \log n)^{1/4}\cdot \frac{\sqrt{\log (2/\delta)}}{\eps}}
\end{align*}
when $k$ is odd.
Moreover, the algorithm runs in time $O(m)\cdot n^{O(k)}\,.$
\end{theorem}

Notice the algorithm runs in time polynomial in the number of queries and the dataset size.

It is known that simply applying the Gaussian mechanism to the input achieves error of the order $O\Paren{\frac{\log (2/\delta)}{\eps^2}}\cdot n^{k}$, which is information theoretically optimal when $m\geq n^{3k/4}\,,$ up to the dependency on $\eps, \delta\,.$
In the more realistic settings of $m\leq n^{3k/4}\,,$ \cref{theorem:k-way-marginals-informal} non-trivially improves over this naive algorithm.
In comparison, existing polynomial time algorithms \cite{dwork2015efficient} could obtain guarantees comparable to \cref{theorem:k-way-marginals-informal} only for \textit{even} $k\,.$
In the odd case, these algorithms yielded an  average query wise squared error larger by an $n^{1/2}/(k\log n)$ multiplicative factor.\footnote{This gap appears in \cite{dwork2015efficient} due to the fact that even order $k$ tensors can be flattened into $k/2\times k/2$ matrices and odd order tensors cannot.} Furthermore, the algorithm of \cite{dwork2015efficient} required constrained convex optimization methods in which random noise is injected at each iteration. 

Disregarding computational complexity, exponential time algorithms \cite{gupta2011privately, hardt2012simple, hardt2010multiplicative} are known to achieve entry-wise error $m\cdot n^{1/4}\,.$ This bound is known to be tight \cite{kasiviswanathan2010price} and it is matched, up to constant factors, by \cref{theorem:k-way-marginals-informal} for $k=2\,.$
Given that the $k$-way marginal queries are the entries of the $k$-th order tensor of the dataset (see \cref{section:techniques}), this information-computation gap appears to be related to the well-known conjectural hardness of computing injective tensor norms of random tensors \cite{bhattiprolu2016sum, hopkins2017power, brennan2020reducibility}.

Despite these strong negative results, it turns out that, for $t$-sparse datasets,\footnote{We say a dataset is $t$-sparse if all vectors in  the dataset have at most $t$ non-zero entries.} one can achieve significantly higher utility than \cref{theorem:k-way-marginals-informal} whenever $t \leq n^{5/6}/\log n\,.$

\begin{theorem}[K-way marginals for sparse-datasets, informal]\label{theorem:k-way-marginals-sparse-informal}
Let $D$ be a dataset of $m$ elements, each a a binary vector in $\set{0,1}^{n}$ with at most $t$ non-zero entries.
There exists an $(\epsilon,\delta)$-differentially private algorithm that,  on input $D\,,k\,,t$, answers all $k$-way marginal queries with expected average query-wise squared error
\begin{align*}
    O\Paren{\frac{\sqrt{\log (2/\delta)}}{\eps}}\cdot \Paren{m\cdot (t^{3/2}/n)^{k}\cdot \sqrt{k \log n/t^2}}
\end{align*}
when $k$ is even,
\begin{align*}
    O\Paren{\frac{\sqrt{\log (2/\delta)}}{\eps}}\cdot \Paren{m\cdot (t^{3/2}/n)^{k}\cdot \sqrt{k \log n}}
\end{align*}
when $k$ is odd.
Moreover, the algorithm runs in time $O(m)\cdot n^{O(k)}$.
\end{theorem}

The error bound of \cref{theorem:k-way-marginals-sparse-informal} may seem counter-intuitive as it becomes vanishing small as $t^{3/2}/n$ decreases. However, this is a consequence of the fact that in such settings most queries must have value $0$.  In particular, the following trivial algorithm achieves vanishing small error for sufficiently small $t^2/n$ ratio: add Gaussian noise and remove all but the largest $m\cdot t^{k}$ entries.

Similarly to the dense settings, the error of \cref{theorem:k-way-marginals-sparse-informal} is related to the existing (conditional) computational lower bounds for certifying sparse injective tensor norms of random tensors \cite{choo2021complexity} (see \cref{section:techniques}). 

%% file: related.tex
\section{Related work}\label{section:related-work}

Differential privacy~\cite{dwork2006calibrating} has emerged in the past decades as the \emph{de facto} privacy standard in the design of private algorithms. There is an ever-growing literature on DP for which we refer to the survey of~\citet{dwork2014algorithmic}. Our work spans many areas of  DP algorithms including similarity and distance approximation and k-way marginal estimation. We now review the most relevant work in each of these areas.  

\paragraph{Similarity and distance approximation}
A related area is that of approximately (and privately) preserving distance and similarity measures among metric data. A highly celebrated result in the non-private literature is that the Johnson-Lindenstrauss (JL) linear projection preserves approximately the distances of Euclidean points~\cite{dasgupta2003elementary}. More recently~\citet{kenthapadi2012privacy} showed that JL projection with additional noise can be used to preserve distances while achieving vector level DP. \citet{blocki2012johnson}~showed instead that, under certain assumptions on the data, the JL projection itself is sufficient to achieve DP (without any additional noise). Subsequent work extended such analysis to private sparse JL projections~\cite{stausholm2021improved}.

Related projection-based methods have found wide application in privacy settings for instance in solving PCA problems~\cite{chaudhuri2013near}, answering statistical queries~\cite{nikolov2023private}, releasing synthetic data~\cite{xu2017dppro}, answering distance based queries~\cite{yang2018differential}, as well as addressing computer vision problems~\cite{zhu2020private}. 

For the related problem of preserving similarity (as opposed to distance) measures, \citet{amuller2020differentially} provided Local DP (LDP) algorithms for Jaccard Similarity estimation in item sets. \citet{yang2017privacy} instead applies JL-based methods for approximating the cosine similarity.  On a related problem, \citet{fernandes2021locality} designed approximate cosine similarity computations in an LDP relaxation known as local extended DP (or $d_\chi$ privacy). 
Our work provides improved bounds for cosine similarity computation over such work for the model of central DP (as opposed to LDP).

On a different direction, 
\citet{alaggan2011private, choiminhash} used cryptographic schemes for cosine similarity based on public-private key encryption and fully-homomorphic encryption. These works require cryptographic assumptions for privacy and focus on the 2-party computations (i.e., two users interested in computing the similarity of their data). \citet{schoppmann2018secure} studied instead secure multi-party computation for performing similarity search in documents.

From an application point of view, \citet{battaglia2021differentially} studied the problem of inferring a distance measure over categorical values in central DP setting using the DILCA method; \citet{xue2016differential, zhang2020privrec} studied recommender systems with differential privacy, while \citet{ding2019privacy, chen2019improved,wang2020searching} presented privacy-aware methods for finding similar items.

\paragraph{k-way marginals}

Marginal tables and contingency tables are a high level synopsis of a multi-dimensional dataset. The problem has been studied in the context of DP for a long time. \citet{gupta2011privately, hardt2012simple, hardt2010multiplicative,kasiviswanathan2010price} showed lowerbounds and algorithms for answering such queries. \citet{chandrasekaran2014faster,dwork2015efficient} focused on efficient algorithms for answering high dimensional marginals with DP. \citet{nikolov2023private} used JL projections to answer 2-way marginals while \citet{cormode2018marginal} focused on a local DP version of the problem. 
From an application point of view marginal tables can be used to produce synthetic data~\cite{li2017publishing}.

Finally a related problem to 2-way marginal is covariance matrix estimation. Several authors~\cite{dong2022differentially,blocki2012johnson,mangoubi2023private} studied the problem in a private setting. 

\paragraph{Sum-of-squares-based algorithms}
Our work is tightly linked to recent advances in robust statistics \cite{hopkins2015tensor,hopkins2018mixture, d2020sparse,kothari2018robust,diakonikolas2019recent, ding2022robust,ding2023node,bakshi2020outlier, bakshi2022robustly, klivans2018efficient,liu2022minimax}). Very recently, the sum-of-squares framework over which we build upon, has found surprising applications in the context of differential privacy \cite{hopkins2023robustness, georgiev2022privacy, hopkins2022efficient} including clustering \cite{chen2023private} and moment estimation \cite{kothari2022private}. Relevant to our techniques are also the results on consistent estimation of
\cite{tsakonas2014convergence, d2021consistent1,d2021consistent2,d2023higher}, see \cref{section:techniques}.

%% file: preliminaries.tex
\section*{Organization}
The rest of the paper is organized as follows. 
We provide a technical overview in \cref{section:techniques}.
In \cref{section:meta-theorem} we prove general guarantees for the perturb-and-project framework. Here we also show how to design practical implementations of the algorithm and what the resulting guarantees are. We then use these results to obtain \cref{theorem:cosine-similarities-release} in \cref{section:distances-release} and to obtain \cref{theorem:k-way-marginals-informal}, \cref{theorem:k-way-marginals-sparse-informal} in \cref{section:k-way-marginals}.
Background notions and definitions used throughout the paper can be found in \cref{section:background}. 

\section*{Notation}
We describe here the notation used throughout the paper and some relevant preliminary notions.
We hide absolute constant multiplicative factors using the standard notation $O(\cdot)\,, \Omega(\cdot)\,, \Theta(\cdot)$.
We denote random variables in \textbf{boldface}. 
We write $o(1), \omega(1)$ for functions tending to zero (resp. infinity) as $n$ grows. We say that an event happens with high probability if this probability is at least $1-o(1)$.
Throughout the paper, when we say "an algorithm runs in time $O(q)$" we mean that the number of basic arithmetic operations involved is $O(q)$.

\paragraph{Vectors, matrices, tensors}We use $\Id_n$ to denote the $n$-by-$n$ dimensional identity matrix, $\bar{0}_n\in \R^n$ to denote  the zero vector.  When the context is clear we drop the subscript. For matrices $A, B \in \R^{n\times n}$ we write $A\sge B$ if $A-B$ is positive semidefinite. 
We denote by $\Norm{M}$ the spectral norm of $M$, by $\Normn{M}$ its nuclear norm, by $\Normf{M}$ its frobenius norm and by $\Normm{X}$ the largest absolute value of any of its entries. We use $M_i$ and $M_{-,i}$ to respectively denote the $i$-th row and column of $M\,.$ 
We denote by $\Paren{\R^{n}}^{\otimes k}$ the set of real-valued order-$k$ tensors. For a $n\times n$ matrix $M$, we denote by $M^{\otimes k}$ the \emph{$t$-fold Kronecker product} $\underbrace{M\otimes M \otimes \cdots \otimes M}_{k\ \text{times}}$.  
We define the \emph{flattening}, or \emph{vectorization}, of $M$ to be the $n^k$-dimensional vector, whose entries are the entries of $M$ appearing in lexicographic order. With a slight abuse of notation we refer to this flattening  with $M$, ambiguities will be clarified form context. 
We denote by $N\Paren{0, \sigma^2}^{n^{\otimes k}}$ the distribution over Gaussian tensors with $n^{k}$ independent entries with standard deviation $ \sigma$.


\paragraph{Sparse vectors and norm} We denote number of non-zero entries of a vector $v\in \R^n$ by $\Norm{v}_0$. Hence a $t$-sparse vector $v$ is a vector satisfying $\Norm{v}_0\leq t\,.$ For an order-$k$ tensor $M$, we define its symmetric innjective $t$-sparse norm as $\max_{v\in \R^n\,, \norm{v}=1\,, \norm{v}_0\leq t}\iprod{M, \tensorpower{v}{k}}\,.$



\paragraph{Sets and projections}
We use calligraphic letters to denote sets.
Given $A\in \R^n$ and a compact convex set $\cS\in \R^n$, we denote by $\Pi_\cS(A)$ the projection of $A$ onto $\cS\,,$ $\Pi_\cS(A):=\arg\min_{X\in \cS}\Snormt{A-X}\,.$
For two sets $\cS,\cS'$ we denote their symmetric difference by $S\triangle S'\,.$ We write $\cG(\cS)$ for the Gaussian complexity of $\cS\,.$

Notions about Gaussian complexity, differential privacy and sum-of-squares are deferred to \cref{section:background}.

%% file: techniques.tex
\section{Techniques}\label{section:techniques}
We present here the main ideas behind our results.

\paragraph{The perturb-and-project framework}  The general problem we consider can be described as follows: given a vector $A\in \R^n,$ and a fixed space $\cS\subseteq \R^n\,,$ we would like to output the projection of $A$ onto $\cS$ while preserving differential privacy with respect to $A$ (for some problem dependent notion of adjacency). 
Our algorithmic approach can be then described as follows.
\begin{algorithm}[ht]
   \caption{Perturb-and-project}
   \label{algorithm:add-noise-and-project}
\begin{algorithmic}
   \STATE {\bfseries Input:}  $\eps,\delta,\Delta>0\,,A\in \R^n\,,$ compact convex set $\cS\subseteq \R^{n}\,.$
    \STATE {\bfseries Output:} $\hat{\mathbf{\Pi}}\in\cS\,.$
    \STATE Let $\mathbf{W}\sim N\Paren{0,\Id_n\cdot \frac{2\log (2/\delta)}{\eps^2}\cdot \Delta^2}\,.$
    \STATE {\bfseries Return:} $\hat{\mathbf{\Pi}}:=\arg\min_{X\in \cS}\Normt{X-A+\mathbf{W}}^2\,.$ 
\end{algorithmic}
\end{algorithm}

Notice that the set $\cS$ is a parameter that is \textit{independent} of the input vector $A$.
It is not hard to show that, for compact convex sets, the strong convexity of the projection guarantees
\begin{align}\label{equation:techniques-gaussian-complexity}
    \E \Snormt{\Pi_\cS(A)-\Pi_\cS(A+\mathbf{W})}\leq O(\tfrac{2\Delta\sqrt{\log (2/\delta)}}{\eps^2})\cdot \cG(\cS)
\end{align}
where $\cG(\cS):=\E_{\mathbf{W}^*\sim N(0,\Id_n)}\Brac{\sup_{X\in \cS}\iprod{X,\mathbf{W}^*}}$ is the Gaussian complexity of $\cS$  and $\Pi_\cS(\cdot)$ denotes the orthogonal projection onto $\cS$. (Related results exist in the literature, see \cref{section:meta-theorem} for a proof and a proper comparison).
The significance of \cref{equation:techniques-gaussian-complexity} is in the implication that what governs the error is the structure of the space $\cS\,.$ Indeed, by Sudakov's minoration \cite{wainwright2019high} the minimal size of any $\alpha$-net over $\cS$ is bounded by $\exp(\cG(\cS)^2/\eps^2)\,.$
If $\cS$ in this set are heavily structured, then the resulting error will be significantly better than naively using the Gaussian mechanism.

\paragraph{A simple application: pair-wise cosine similarities}
As a concrete example, let $\set{v_1,\ldots,v_n}$ be a set of vectors in $\set{\pm 1/\sqrt{m}}^m$  (this assumption is only for simplicity of the exposition) and let $V\in \R^{n \times m}$ be the matrix with rows $V_i=v_i$.
The problem of privately releasing the cosine similarities between these vectors is equivalent to the problem of privately realising $\dyad{V}\,.$ 
Notice that this matrix falls in the set 
\begin{align}\label{equation:techniques-cosine-similarity}
    \cS:=\Set{X\in \R^{n\times n}\suchthat X\sge 0\,, X_{ii}\leq 1\,, \forall i}\,.
\end{align}
In other words, this task adheres perfectly to the perturb-and-project framework with $A=\dyad{V}\,.$
Hence we can apply the  Gaussian mechanism and then project onto the set in~\cref{equation:techniques-cosine-similarity}.
Since by Holder's inequality for the nuclear and the spectral norm,
\begin{align*}
    \E_{\mathbf{W}\sim N(0,\Id_n)^{n\times n}}\Brac{\sup_{X\in \cS}\iprod{X,\mathbf{W}}}&\leq  \E\Brac{\Normn{X}\Norm{\mathbf{W}}}\\
    &\leq n\cdot 2\sqrt{n}\cdot \sigma\,,
\end{align*}
we immediately recover the guarantees of \cref{theorem:cosine-similarities-release} for the appropriate choice of $\sigma\,.$

\paragraph{Tight relaxations and sum-of-squares}
In the above examples, the projection onto the desired set $\cS$ could be computed efficiently.
However, this is not the case in general.
For example, if $A$ is (the flattening of) a rank-$m$ $k$-th order symmetric tensor, it stands to reason one may want to project the perturbed input onto the set of rank-$m$ $k$-order symmetric tensors\footnote{The careful reader may argue that this set is not convex.
However, notice we may obtain comparable guarantees by projecting onto the convex set of $k$-th order tensors that are expectations of distributions over rank-$m$ tensors.}
\begin{align}\label{equation:techniques-k-way-marginals}
    \Set{X\in (\R^n)^{\otimes k}\suchthat \sum_{i\in [m]}\tensorpower{v_i}{k}=X\,, \Normm{X}\leq 1}\,.
\end{align}
Unfortunately, carrying out such a projection is (conjecturally) hard even on average \cite{bhattiprolu2016sum, hopkins2017power, brennan2020reducibility}.
To overcome this issue, one may replace $\cS$ with a superset $\cS'$ over which we can efficiently project.
The hope is that, if $\cS'$ is not too "large", then $\cG(\cS')$ is close to $\cG(\cS)$ and therefore the error is not significantly larger. Following \cite{d2023higher} our idea  to find a tight set $\cS'\,,$ is to use the natural degree-$O(1)$ sum-of-squares relaxation of $\cS\,.$

\paragraph{An advanced application: k-way marginal queries}
The problem of releasing $k$-way marginal queries can indeed be recast as the above example (see \cref{section:k-way-marginals}).
Given a set $\set{e_1,\ldots,e_n}$ in $\set{0,1}^m\,,$ privately output the tensor $\tfrac{1}{m}\sum_i \tensorpower{e_i}{k}\,.$ Here, adjacent datasets are usually defined as datasets differing in one vector. 
The even $k$ case is easy as one can always flatten the tensor into a $n^{k/2}$-by-$n^{k/2}$ matrix. Indeed this is the approach of \cite{dwork2015efficient}. Hence we discuss the odd case.
Ideally, one would like now to project the perturbed input exactly onto the set of rank-$m$ tensors in \cref{equation:techniques-k-way-marginals}.
However, to run the algorithm efficiently, we replace this set with its sum of squares relaxation:
\begin{align*}
    \Set{
    X\in (\R^n)^{\otimes k}\,|\,\exists \textnormal{ deg-$O(k)$ }\mu \textnormal{ over }\cQ \textnormal{ s.t. } \tilde{\E}_\mu\brac{\sum_{i\in [m]}\tensorpower{v_i}{k}}=X }\,,
\end{align*}
where $\cQ$ is the set of polynomial inequalities over vector variables $v_1,\ldots,v_n$
\begin{align*}
    \cQ:=\Set{\Normm{\sum_i \tensorpower{v_i}{k}}\leq 1}\,.
\end{align*}
The Gaussian complexity of $\cS$ is bounded by $O(\sqrt{k\log n})\,,$ in contrast for this set $\cS'$ 
\begin{align*}
    \cG(\cS')&=\E_{\mathbf{W}\sim N(0,\Id_n)}\Brac{\sup_{X\in \cS'}\iprod{X,\mathbf{W}}}\\
    &=\E_{\mathbf{W}\sim N(0,\Id_n)}\Brac{\sup_{\textnormal{deg-O(1) }\mu \textnormal{ over }\cQ}\iprod{\tilde{\E}_\mu\Brac{\sum_i \tensorpower{v_i}{k}},\mathbf{W}}}\,.
\end{align*}
This last term is the minimum value of any degree-$O(1)$ sum-of-squares certificate of the injective tensor norm of Gaussian tensors, which is known to be $O(n^{k/2})$ up to multiplicative logarithmic factors \cite{hopkins2015tensor, bhattiprolu2016sum}.
This approach yields the guarantees of \cref{theorem:k-way-marginals-informal}

\paragraph{Sparse k-way marginal queries}
The aforementioned ideas can be pushed even further.
For instance, real world datasets over which $k$-way marginal queries are relevant are often sparse. For instance, consider the cause of a store interested in estimating the number of users that have purchased any specific basket of item. It is natural to assume that no user has purchased a large fraction of the whole inventory.  Therefore, it is natural to ask whether one could improve over the above guarantees when each vector $e_i$ in the datasets has at most $t$ non-zero entries: $e_i \in \Set{v\in \R^n\suchthat \Norm{v}_0\leq t}\,.$ Here one would like to project onto the intersection between this set and the one in \cref{equation:techniques-k-way-marginals}. The Gaussian complexity of this intersection can be verified to be $O(\sqrt{t\log n})$ \cite{choo2021complexity} but again, projecting onto this set is computationally hard.
Our improvement in these settings come from studying the sum-of-squares relaxations of this set (see \cref{section:sos-sparse-certificates}).  Specifically, we introduce \textit{novel} sum-of-squares certificates for sparse injective tensor norms of Gaussian tensors. Our bounds match existing lower bounds (against the class of algorithms captured by low degree polynomials) up to logarithms \cite{choo2021complexity}.

\paragraph{Practical perturb-and-project via alternating projections}
The algorithms discussed so far show how effective the perturb-and-project framework can be.
However, even though these projections can be computed in polynomial time, at first sight comparable results appear unattainable in practice given that the computational budget required to naively implement these projections is too large for real-world applications.
This warrant the question of whether comparable guarantees could be achieved in practice. It turns out that this is the case to a large extent.

A long line of work---see \cite{bauschke1993convergence, sakai1995strong} and references therein---initiated by \cite{von1949rings} showed how the orthogonal projection onto a  set that is the intersection of multiple compact convex sets $\cS={\bigcap}_{\ell\in [q]}\cS_\ell$ can be computed by repeatedly projecting onto each of these sets and averaging the resulting projections.
Remarkably, the convergence of this iterative algorithm is linear in the number of iterations, implying that whenever $\Snormt{A}\leq n^{O(1)}$ a logarithmic number of steps suffices.
In our settings, this amounts to replacing step $2$ in \cref{algorithm:add-noise-and-project}  as described (\cref{algorithm:add-noise-and-alternate-projections}).
\begin{algorithm}[ht]
   \caption{Perturb-and-alternately-project}
   \label{algorithm:add-noise-and-alternate-projections}
\begin{algorithmic}
   \STATE {\bfseries Input:}  $\eps,\delta, \Delta,t>0\,,A\in \R^n$ compact convex sets $\cS\,, \cS_1\,,\ldots\,,\cS_q\subseteq \R^{n}$ with $\bigcap_{\ell \in [q]}\cS_\ell=\cS\,.$
    \STATE {\bfseries Output:} $\hat{\mathbf{\Pi}}\,.$
    \STATE Let $\mathbf X_0=A+\mathbf W$ for $\mathbf{W}\sim N\Paren{0,\Id_n\cdot \frac{2\log (2/\delta)}{\eps^2}\cdot \Delta^2}\,.$
    \FOR{$i=1$ {\bfseries to} $t$}
    \STATE Compute $\mathbf{X}_{i}=\E_{\mathbf{\ell} \sim [q]}\arg\min_{X\in \cS_{\mathbf \ell}}\Snormt{X-\mathbf X_{i-1}}$
    \ENDFOR
    \STATE {\bfseries Return} $\hat{\mathbf{\Pi}}:=\mathbf X_t\,.$
\end{algorithmic}
\end{algorithm}

At each iteration $i\,,$
\cref{algorithm:add-noise-and-alternate-projections} projects the current vector $\mathbf{X}_{i-1}$ to \textit{each}  of the sets $\cS_1,\ldots,\cS_q$ and then computes the average.
A single iteration of \cref{algorithm:add-noise-and-alternate-projections} can be significantly faster than \cref{algorithm:add-noise-and-project} as well as easier to implement. This is because each of the projections may be fast to compute by itself. 
Furthermore, since by linear convergence few iterations suffice, the whole algorithm turns out to be faster than \cref{algorithm:add-noise-and-project}.

As a concrete example, observe that the set of \cref{equation:techniques-cosine-similarity}, is the intersection of\footnote{Technically speaking we would like all sets considered to be explicitly bounded. This can be easily achieved by replacing the positive semidefinite cone with its intersection with a large enough Euclidean ball.} $\set{X\in \R^{n\times n}\,|, X\sge 0}\,,$ and $\Set{X\in \R^{n\times n}\,|\, \Normm{X}\leq 1}\,.$ 
Hence we can compute the output of our algorithm repeatedly projecting onto these two sets and averaging. The projection onto the first set can be obtained zeroing out negative eigenvalues. The latter projection amounts to clipping entries that are larger than $1$ in absolute values, a typical subroutine of practical differentially private algorithms!

That is, not only the perturb-and-alternately-project framework can yield practical algorithms, it also gives an alternative perspective on typical subroutines used  in the context of privacy.

\begin{remark}[On the error guarantees of \cref{theorem:cosine-similarities-release}]\label{remark:distance-releases-optimality}
The task of computing $2$-way marginal queries under differential privacy is related to the task of privately computing pair-wise cosine similarities.
One can see the vectors over which we compute the cosine similarities as the Gram vectors of the $2$-way marginal queries matrix.
Indeed, for certain family of vectors, the set over which we perform the projection in the perturb-and-project framework is the same. By optimality of the algorithm for $k$-way marginal queries \cite{kasiviswanathan2010price}, this suggests that the guarantees of \cref{theorem:cosine-similarities-release} may be optimal among algorithms captured by the perturb-and-project framework.
\end{remark}


%% file: meta-theorem.tex
\section{Guarantees of the perturb-and-project paradigm for differential privacy}\label{section:meta-theorem}

We present here a general statement about the privacy and utility performance of the \textit{perturb-and-project} framework defined in \cref{algorithm:add-noise-and-project}. 

Both the privacy and utility guarantees of \cref{algorithm:add-noise-and-project} can be conveniently expressed in terms of $\epsilon\,,\delta$ and the convex set $\cS\,.$ 
In what follows we say  $A,A'\in \R^n$ are \textit{adjacent} if $\Normt{A-A'}\leq \Delta\,,$ notice this applies to higher order tensors via flattening.
For a convex set $\cS$ and $A\in \R^n$ we define $\Pi_\cS(A):=\arg\min_{X\in \cS}\Snormt{A-X}\,.$

\begin{theorem}[Guarantees of perturb-and-project]\label{theorem:guarantees-add-noise-and-project}
Let $\cS\subseteq \R^n$ be a compact convex set and let $\epsilon,\delta>0$.
Then, on input $ A\in \R^n, \eps,\delta$, \cref{algorithm:add-noise-and-project} returns $\hat{\mathbf{\Pi}}\in \cS$ satisfying
\begin{align*}
    \E \Snormt{\hat{\mathbf{\Pi}}-\Pi_\cS(A)}\leq \frac{8\sqrt{\log(2/\delta)}}{3\eps}\cdot \Delta\cdot\cG(\cS)\,.
\end{align*}
Moreover, the algorithm is $(\epsilon, \delta)$-differentially private.
\end{theorem}

\cref{theorem:guarantees-add-noise-and-project} states that for any given set of privacy parameters $(\eps, \delta)\,,$ the error guarantees of \cref{algorithm:add-noise-and-project} are governed by the metric entropy of the set considered.\footnote{Recall, by Sudakov's minoration, we can bound the metric entropy of a set with an exponential function of its squared Gaussian complexity.} Hence more structured sets yields stronger guarantees. This phenomenon introduces  a trade-off between the computational complexity required by the projection and the accuracy of the estimate.

It is important to remark that similar results appeared in the risk minimization literature \cite{chaudhuri2008privacy, chaudhuri2011differentially, jain2012differentially, kifer2012private, thakurta2013differentially, duchi2013local, jain2014near, bassily2014private, ullman2015private, chourasia2021differential, nikolov2013geometry}. 
The perturb-and-project framework was first studied in \cite{nikolov2013geometry}.
In \cite{talwar2014private}, the authors studied objective perturbations, bounding the error in terms of Gaussian complexity of the solution space the change  in the objective value for strongly convex functions.
Furthermore, \cite{dwork2015efficient} also obtained comparable guarantees for the Frank-Wolfe algorithm, in which noise is added at each iteration.\footnote{Indeed, the argument in \cite{dwork2015efficient} is closely related to the proof of \cref{theorem:guarantees-add-noise-and-project}.}
This difference leads to an important algorithmic consequence: since in \cref{algorithm:add-noise-and-project} both the privacy and error guarantees do not depend on the particular steps involved in the computation of $\hat{\bm \Pi}$, different techniques may be used to effectively carry out the computation of \cref{algorithm:add-noise-and-project} (see \cref{corollary:guarantees-add-noise-and-alternate-projections} and the related discussion).


The privacy guarantees follows from the definition of adjacent inputs and the use of the Gaussian mechanism (\cref{lemma:gaussian-mechanism}).
The utility guarantees follows from an argument about the stability of projections onto compact convex sets, as shown below.

\begin{lemma}[Stability of projections]\label{lemma:stability-of-projections}
   Let $\cS\subseteq \R^n$ be a compact convex set and for any $A\in \R$ let 
    \begin{align*}
        \Pi_\cS(A):=\arg\min_{X\in \cS}\Snormt{X-A}\,.
    \end{align*}
    Then 
    \begin{align*}
        \E_{\mathbf{W}\sim N(0,\Id_n)}\Snormt{\Pi_\cS(A+\mathbf{W})-\Pi_{\cS}(A)}\leq \frac{4}{3}\cG(\cS)\,.
    \end{align*}
    \begin{proof}
        The statement follows by strong convexity of the underlying function.
        Indeed we have (we drop the subscript $\cS$ for simplicity)
        \begin{align*}
            \Snormt{A-\Pi(A)}&\geq \Snormt{A-\Pi(A+\mathbf{W})}+\iprod{\nabla \Snormt{A-\Pi(A+\mathbf{W})}, \Pi(A)-\Pi(A+\mathbf{W})}+\frac{1}{2}\Snormt{\Pi(A+\mathbf{W})-\Pi(A)}
        \end{align*}
        and
        \begin{align*}
            \Snormt{A+\mathbf{W}-\Pi(A+\mathbf{W})}
            &\geq \Snormt{A+\mathbf{W}-\Pi(A)}\\
            &+\iprod{\nabla \Snormt{A+\mathbf{W}-\Pi(A)}, \Pi(A+\mathbf{W})-\Pi(A)}+\frac{1}{2}\Snormt{\Pi(A+\mathbf{W})-\Pi(A)}\,.
        \end{align*}
        Putting the two inequalities together and using the definition  of $\Pi(A+\mathbf{W})\,, \Pi(A)$
        \begin{align*}
            \Snormt{\Pi(A+\mathbf{W})-\Pi(A)}
            &\leq \iprod{\nabla \Snormt{A-\Pi(A+\mathbf{W})}, \Pi(A)-\Pi(A+\mathbf{W})}+ \iprod{\nabla \Snormt{A+\mathbf{W}-\Pi(A)}, \Pi(A+\mathbf{W})-\Pi(A)}\\
            &=-2\Snormt{\Pi(A)-\Pi(A+\mathbf{W})} + 2\iprod{\mathbf{W}, \Pi(A+\mathbf{W})-\Pi(A)}\,.
        \end{align*}
        Rearranging and taking expectations
        \begin{align*}
            \Snormt{\Pi(A+\mathbf{W})-\Pi(A)}
            &\leq \frac{2}{3}\E\iprod{\mathbf{W}, \Pi(A+\mathbf{W})-\Pi(A)}\leq \frac{4}{3}\E\cG(\cS)\,.
        \end{align*}
    \end{proof}
\end{lemma}

\subsection{Practical perturb-and-project algorithms}\label{section:practical-perturb-and-project}

As mentioned in the previous sections, a crucial consequence of \cref{theorem:guarantees-add-noise-and-project} is that the projection steps of \cref{algorithm:add-noise-and-project} can be broken down into a sequence of simpler projections as in \cref{algorithm:add-noise-and-alternate-projections}.
We study here the guarantees of this algorithm.

\begin{corollary}[Guarantees of the perturb-and-alternately-project]\label{corollary:guarantees-add-noise-and-alternate-projections}
Let $\eps,\delta\geq 0\,.$
Let $$\cB_C:=\Set{X\in \R^{n}\suchthat \Snormt{X}\leq n^C}\subseteq \R^n$$ and let $\cS_1,\ldots,\cS_n\subseteq\R^n$ be compact convex sets with intersection $\bigcap_{\ell \in [q]}\cS_\ell=\cS\subseteq \cB_C$  for some $C\geq 0\,.$
Then, on input $A\in \R^n,\eps,\delta, t \geq O(C \log n)\,, \Delta\leq n^{O(1)}$, \cref{algorithm:add-noise-and-alternate-projections} returns $\hat{\mathbf{\Pi}}_t$ satisfying
\begin{align*}
    \E \Snormt{\hat{\mathbf{\Pi}}_t-\Pi_\cS(A)}\leq& \frac{16\sqrt{\log(2/\delta)}}{3\eps}\cdot\Delta\cdot\Paren{\cG(\cS)+ 2^{-t}\Snormt{A}+ n^{-\Omega(1)}}\,.
\end{align*}
Moreover, the algorithm is $(\epsilon, \delta)$-differentially private.
\end{corollary}

The proof of \cref{corollary:guarantees-add-noise-and-alternate-projections} rely on a result in the theory of alternating projections \cite{bauschke1993convergence} (see \cite{sakai1995strong} and references therein for variations of \cref{algorithm:add-noise-and-alternate-projections}).

\begin{theorem}[Linear convergence of alternating projections, \cite{bauschke1993convergence}]\label{theorem:convergence-alternating-projections}
    Let $\cS_1,\ldots,\cS_q\subseteq\R^n$ be compact convex sets with intersection $\cS=\underset{i\in [q]}{\cap}\cS_i\,.$ For $x\in \R^n\,,$ let $\Set{\Pi_i(x)}_{t\geq 1}$ be the sequence with $\Pi_0(x)=x$ and $\Pi_t(x)=\E_{\mathbf{i}\overset{u.a.r}{\sim} [q]}\Pi_{S_{\mathbf{i}}}(\Pi_{t-1}(x))\,.$ Then, there exists a constant $c\in (0,1)\,,$ depending only on $\cS_1,\ldots, \cS_q\,,$ such that
    \begin{align*}
        \Normt{\Pi_\cS(x)-\Pi_t(x)} \leq c^t\Normt{\Pi_\cS(x)-x}\,.
    \end{align*}
\end{theorem}

\cref{theorem:convergence-alternating-projections} states that the output of \cref{algorithm:add-noise-and-alternate-projections} converges to the output of \cref{algorithm:add-noise-and-project} linearly in the number of iterations.
We are ready to prove \cref{corollary:guarantees-add-noise-and-alternate-projections}.

\begin{proof}[Proof of \cref{corollary:guarantees-add-noise-and-alternate-projections}]
For a matrix $M\in \R^{n\times n}\,,$  let  ${\Pi}_t(M)$ be the output of step 2 of \cref{algorithm:add-noise-and-alternate-projections} with $t$ iterations and $\Pi(M)$ the true projection of $M$ onto $\cS\,.$
On input $A$, we can bound the expected squared error of \cref{algorithm:add-noise-and-alternate-projections} as follows:
\begin{align*}
    \E_{\mathbf{W}}\Snormt{\Pi_t(A+\mathbf{W})-\Pi(A)}
    &\leq 2\E_{\mathbf{W}}\Snormt{\Pi_t(A+\mathbf{W})-\Pi(A+\mathbf{W})}+ 2\E_{\mathbf{W}}\Snormt{\Pi(A+\mathbf{W})-\Pi(A)}
\end{align*}
where we used Cauchy-Schwarz.
By \cref{theorem:convergence-alternating-projections}
\begin{align*}
    \E_{\mathbf{W}}\Snormt{\Pi_t(A+\mathbf{W})-\Pi(A+\mathbf{W})}
    &\leq \E_{\mathbf{W}} c^t\Snormt{A+\mathbf{W}- \Pi(A+\mathbf{W})} \\
    &\leq 2c^t\Paren{\Snormt{A}+\Delta^2 \cdot n^2 + n^C}\\
    &\leq 2c^t \Snormt{A}+ 1/n^{-10}\,,
\end{align*}
where in the last step we used the fact $t\geq (12+12C)\log_c n\,.$
Combining this with \cref{theorem:guarantees-add-noise-and-project} the result follows as desired.
\end{proof}

%% file: distances-release.tex
\section{Cosine similarity release}\label{section:distances-release}

We introduce here our differentially private algorithm to compute the pair-wise similarity between a set of vectors and prove \cref{theorem:cosine-similarities-release}.

\begin{algorithm}[ht]
   \caption{Private pair-wise cosine similarities}
   \label{algorithm:cosine-similarity}
\begin{algorithmic}
   \STATE {\bfseries Input:}  $\eps,\delta,\Delta>0\,,V\in \R^{n\times m}\,.$
    \STATE {\bfseries Output:} $\hat{\mathbf{X}}\in \R^{n\times n}\,.$
    \STATE Run \cref{algorithm:add-noise-and-project} with input $\dyad{V}$, parameters $\eps,\delta,\Delta$ and $\cS=\Set{X\in \R^{n\times n}\suchthat X\sge 0\,, X_{ii}\leq 1}\,.$
\end{algorithmic}
\end{algorithm}

Equipped with \cref{algorithm:cosine-similarity} we are ready to prove \cref{theorem:cosine-similarities-release}.

\begin{proof}[Proof of \cref{theorem:cosine-similarities-release}]
    Notice that $V\transpose{V}\in \R^{n\times n}$ is a positive semidefinite matrix satisfying $\Normm{\dyad{V}}\leq 1\,.$
    Moreover, by definition of adjacency, for any  $V'\in \R^{n\times m}$ adjacent to $V$  we have
    \begin{align*}
        \Delta^2
        \geq \Normf{\dyad{V}-\dyad{V'}}^2\,.
    \end{align*}
    Privacy of \cref{algorithm:cosine-similarity} then follows by \cref{theorem:guarantees-add-noise-and-project}.
    Concerning utility, notice that by Holder's inequality,
    \begin{align*}
        \cG(\cS)&=\E_{\mathbf{W}^*\sim N(0,1)^{n\times n}}\sup_{X\in \cS}\iprod{X, \mathbf{W}^*}\leq \Normn{X}\E\Norm{\mathbf{W}^*}\leq n\cdot 2\sqrt{n}\,.
    \end{align*}
    The desired error bound follows again by \cref{theorem:guarantees-add-noise-and-project}.
\end{proof}

\begin{remark}[On using \cref{algorithm:add-noise-and-alternate-projections} for pair-wise cosine similarity]\label{remark:cosine-similarity}
Notice that in practice it is hard to implement the projection used in the last step of \cref{algorithm:add-noise-and-project}.
By \cref{corollary:guarantees-add-noise-and-alternate-projections} we may replace this step with an application of \cref{algorithm:add-noise-and-alternate-projections} with $\cS_1:=\Set{X\in \R^{n\times n}\suchthat \Normf{X}\leq n}$ and $\cS_2:=\Set{X\in \R^{n\times n}\suchthat \Normm{X}\leq 1}\,.$
Indeed it always holds that $\dyad{V}\in \cS_1\cap \cS_2\,.$
Now the projection of a matrix $X\in \R^{n\times n}$ onto $\cS_1$ can be obtained removing negative eigenvalues and rescaling the positive eigenvalues $\lambda_1,\ldots,\lambda_n\geq 0$ so that $\sum_i \lambda_i^2\leq n\,.$ The projection onto $\cS_2$ can be obtained by clipping entries to $1$ in absolute value. Moreover, by \cref{corollary:guarantees-add-noise-and-alternate-projections} a logarithmic number of iterations suffices.
The algorithm then is the following.
\begin{algorithm}[ht]
   \caption{Practical pair-wise cosine similarities}
   \label{algorithm:practical-cosine-similarity}
\begin{algorithmic}
   \STATE {\bfseries Input:}  $\eps,\delta,\Delta,t>0\,,V\in \R^{n\times m}\,.$
    \STATE {\bfseries Output:} $\hat{\mathbf{X}}\in \R^{n\times n}\,.$
    \STATE Let $X^{(0)}=\dyad{V}+\mathbf{W}$, where $\mathbf{W}\sim N(0,\tfrac{\delta^2\cdot \log (2/\delta)}{\eps^2})^{n\times n}\,.$
    \FOR{$i=1$ to $t$}
    \STATE Compute $\mathbf X^{(i)}=\tfrac{1}{2}\Pi_{\cS_1}(\mathbf X^{(i-1)})+\tfrac{1}{2}\Pi_{\cS_2}(\mathbf X^{(i-1)})\,.$
    \ENDFOR
    \STATE {\bfseries Return:} $\mathbf X^{(t)}\,.$
\end{algorithmic}
\end{algorithm}
\end{remark}

%% file: k-way-marginals.tex
\section{K-way-marginal queries and sparse tensor certificates}\label{section:k-way-marginals}
In this section we combine \cref{theorem:guarantees-add-noise-and-project} with novel sum-of-squares certificates for injective tensor norms to obtain tight results for $k$-way marginal queries. Before formally stating the theorems we introduce some definitions.

Let $[n]$ denote a set of $n$ features. A user can be represented by a binary vector  $e\in\Set{0,1}^n\,.$ There are at most $2^n$ distinct potential users with features in $[n]$. 
We denote the set of all possible $2^n$ distinct potential users by $\cE:$ and the multiplicity of $e\in \cE$ in a dataset $D$ by $m_{D}(e)$.
Hence, a dataset $D$ can also be described by the $2^n$ dimensional vector $m(D)$ with entries $m(D)_e=m_D(e)$ for each $e\in \cE\,.$ 
When the context is clear we drop the subscript. We denote the set of all possible datasets over $m$ users as $\cD_{m,n}\,,$ and by $\cM_m$ the set of all possible vectors $m(D)$ with $D\in\cD_{m,n}$. Notice that $\card{\cD_{m,n}}=\card{\cM_n}\,.$ We are interested in differential privacy with respect to the following natural notion of adjacent datasets.

\begin{definition}[Adjacent datasets]\label{definition:k-way-adjacent-datasets}
Two datasets $D, D'\in\cD_{m,n}$ are said to be \textit{adjacent} if 
\begin{align}\label{equation:kway-sensitivity}
    \Normo{m(D)-m(D')}\leq 2\,.
\end{align}
In other words, $D'$ can be obtained from $D$ replacing one user in the dataset.
\end{definition}

We will formally study so-called \textit{k-wise parity} queries. Indeed, as shown in  \cite{barak2007privacy, dwork2015efficient}, k-way parities form an orthonormal base of all k-way marginals. In other words, it suffices to reconstruct k-wise parity queries to answer all k-way marginal queries.

\begin{definition}[k-wise parity query]\label{definition:k-wise-parity-query}
    For $D\in \cD_{m,n}\,,$ a k-wise parity query is specified by some $\alpha\subseteq [n]$  and given by
    \begin{align*}
        \sum_{e\in \cE} m_D(e)\cdot \underset{i\in \alpha}{\prod}e_i\,.
    \end{align*}
\end{definition}
In particular, \cref{definition:k-wise-parity-query} implies that the tensor 
\begin{align*}
    \sum_{e\in \cE}m_D(e)\tensorpower{e}{k}
\end{align*}
contains \textit{all} $k'$-wise parity queries for any $k'\leq k\,.$
Hence, reconstructing this tensor suffices to answer all k-wise parity queries. 
For sparse datasets, we can capture the sparsity through the following definition.

\begin{definition}[$t$-sparse dataset]\label{definition:sparse-dataset}
Let $\cE_t:=\set{e\in \cE \,|\, \norm{e}_0\leq t}\subseteq \cE$. That is, the set of $t$-sparse users. A dataset $D\in \cD_{m,n}$ is said to be $t$-sparse if $m_D(e)=0$ for all $e\in \cE\setminus \cE_t\,.$ We write $\cD_{m,n,t}\subseteq\cD_{m,n}$ for the subset of $t$-sparse datasets. 
\end{definition}

We are ready to state the main theorem of the section, which implies \cref{theorem:k-way-marginals-informal} and \cref{theorem:k-way-marginals-sparse-informal}. 

\begin{theorem}\label{theorem:k-way-marginals-formal}
    Let $D\in \cD_{m,n,t}\,.$ There exists an $(\epsilon,\delta)$-differentially private algorithm that, on input $D\,, k\,,t\,,$ returns a tensor $\mathbf{T}\in \Paren{\R^n}^{\tensor k}$ satisfying
    \begin{align*}
        \E \Snormt{\mathbf{T}-\sum_{e\in \cE}m_D(e)\tensorpower{e}{k}}\leq O\Paren{ \frac{\sqrt{\log(2/\delta)}}{\eps}\cdot  \card{D}}\cdot \tau\,,
    \end{align*}
    where  if $k$ is even $\tau :=  \min \Set{n^{5k/4}\,, t^{3k/2}\sqrt{k\log n/t^2}}\,,$
    if $k$ is odd $\tau:=  \min \Set{ \Paren{n^{5k} k \log n}^{\frac{1}{4}}\,, (t^{3k}k\log n)^{\frac{1}{2}}}\,.$
    
    Moreover, the algorithm runs in time $m\cdot n^{O(k)}\,.$
\end{theorem}

\cref{theorem:k-way-marginals-formal} immediately implies \cref{theorem:k-way-marginals-informal}, \cref{theorem:k-way-marginals-sparse-informal} by taking the entry-wise average.
The proof of \cref{theorem:k-way-marginals-formal} requires novel sum-of-squares certificates for sparse tensor norms. We first focus on those.

\subsection{Sparse tensor norms certificates via sum-of-squares}\label{section:sos-sparse-certificates}
We make use of the following system of polynomial inequalities in $n$ dimensional vector variables $x,s,y$:

\begin{equation}\label{equation:sparse-tpca-program}\tag{\(\cP_{t}\)}
	\cP_{t}\colon=
	\left \{
	\begin{aligned}
		&\text{unit norm:}& \Snorm{x}\leq 1&\\
		&\text{indicators:}&s_i^2=s_i\\
		&& x_i\cdot s_i=x_i\\
		&\text{sparsity:}&\sum_i s_i\leq t\\
		&\text{absolute value:}& x_i\leq y_i\\
		&& -x_i\leq y_i\\
		&&x_i^2=y_i^2\\
	\end{aligned}
	\right \}
\end{equation} 

\ref{equation:sparse-tpca-program} is the natural sum-of-squares relaxation for the set of $t$-sparse vectors over the unit ball, with the addition of the vector variable $y$ meant to capture the absolute values of entries in $x\,.$
Indeed, it is easy to see that for any $p$-sparse unit vector $z$ in $\R^n$ there exists a corresponding solution $(x,s,y)$ to $\cP_t$ with $x=z$. 
We are ready to state the main theorem of the section.

\begin{theorem}\label{theorem:sos-sparse-certificates}
    Let $\mathbf{W}\in \Paren{\R^n}^{\tensor k}$ be a tensor with i.i.d entries sampled from $N(0,1)\,.$ 
    Let  $\Omega_{k,n}$ be the set of degree-$k$ pseudo-distributions $\mu$  satisfying \ref{equation:sparse-tpca-program}, let 
    \begin{align*}
    &\cS_{4k-4,n}:=\\&\Set{X\in \Paren{\R^n}^{\tensor k}\suchthat \exists\mu\in \Omega_{4k-4,n}\textnormal{  with } \tilde{\E}_\mu \Brac{\tensorpower{x}{k}}=X}\,.
    \end{align*}
    
    Then, it holds
    \begin{itemize}
        \item if $k$ is even $\cG(\cS_{4k-4,n})\leq O\Paren{t^{k/2}\cdot \sqrt{k\log n/t^2}}\,,$
        \item if $k$ is odd $\cG(\cS_{4k-4,n})\leq O\Paren{t^{k/2}\cdot \sqrt{k\log n}}\,.$
    \end{itemize}
\end{theorem}

As already discussed, the even case follows directly from known sum-of-squares bounds for sparse vector norms. The odd case relies crucially on the next lemma.



\begin{lemma}\label{lemma:sos-sparse-tensor-certificate}
Let $A\in \Paren{\R^n}^{\tensor k}\,.$
Then for any degree $4k-4$ pseudo-distribution satisfying \ref{equation:sparse-tpca-program}
\begin{align*}
    \tilde{\E}\iprod{A, \tensorpower{x}{k}} \leq \Normi{A}\cdot t^{\frac{k}{2}}\,.
\end{align*}
\begin{proof}
    For a $n$-dimensional vector $v$ and a multi-set $\alpha\in [n]$ we write $v^{\alpha}=\underset{i\in \alpha}{\prod} v^\alpha\,.$
We have
\begin{align*}
    \tilde{\E}\iprod{A, \tensorpower{x}{k}}&=\tilde{\E}\sum_{\alpha \in [n]^{k}}A_\alpha x^\alpha\cdot s^\alpha\\
    &\leq \tilde{\E}\sum_{\alpha \in [n]^{k}}\Normi{A} y^\alpha\cdot s^\alpha\\
    &\leq \Normi{A}\cdot \Paren{\tilde{\E}\sum_{\alpha \in [n]^{k}} y^{2\alpha}}^{1/2}\cdot \Paren{\tilde{\E}\sum_{\alpha \in [n]^{k}} s^{2\alpha}}^{1/2}\\
    &=\Normi{A}\cdot \Paren{\tilde{\E}\sum_{\alpha \in [n]^{k}} x^{2\alpha}}^{1/2}\cdot \Paren{\tilde{\E}\sum_{\alpha \in [n]^{k}} s^{2\alpha}}^{1/2}\,,
\end{align*}
using \ref{equation:sparse-tpca-program} and Cauchy-Schwarz.
Then, since \ref{equation:sparse-tpca-program} $\sststile{2q}{x}\set{\Normt{x}^{2q}\leq 1}$ and \ref{equation:sparse-tpca-program} $\sststile{2q}{s}\set{\Normt{s}^{2q}\leq t^q}$ we may conclude
\begin{align*}
    \tilde{\E}&\iprod{A, \tensorpower{x}{k}}= \Normi{A}\cdot \Paren{\tilde{\E}\Brac{\Normt{x}^{2k}}\cdot \tilde{\E}\Brac{\Normt{s}^{2k}}}^{1/2}\leq \Normi{A}\cdot t^{k/2}
\end{align*}
as desired.
\end{proof}
\end{lemma}

We can now obtain \cref{theorem:sos-sparse-certificates}.

\begin{proof}[Proof of \cref{theorem:sos-sparse-certificates}]
    First notice that by standard concentration arguments,  $\E\Normi{\mathbf{W}}\leq O\Paren{\sqrt{k\log n}}\,.$
    Now, consider the settings with even $k\,.$ Using the natural matrix flattening of $\tilde{\E}\tensorpower{x}{k}$ and $\mathbf{W}$, the statement follows by \cref{fact:sos-sparse-certificate}. For odd $k$ the argument is an immediate corollary of \cref{lemma:sos-sparse-tensor-certificate}.
\end{proof}

\subsection{Putting things together}
We are ready to prove \cref{theorem:k-way-marginals-formal}.
\begin{proof}[Proof of \cref{theorem:k-way-marginals-formal}]
    By closure under post-processing of differential privacy, we need to add noise once and may then perform the two projections without any additional privacy loss.
    So consider first the dense case and let $T^*:= \frac{1}{\card{D}}\sum_{e\in \cE}m_D(e)\tensorpower{(e/\sqrt{n})}{k} \,.$ This normalization is chosen so that the $\ell_2^2$-sensitivity is $4/\card{D}^2$ and $\Snormt{T^*}=1\,.$  
    Let $\Omega_{k,n}$ be the set of degree-$k$ pseudo-distribution over $\Set{\Snorm{x}\leq 1}\,.$ 
    We use \cref{algorithm:add-noise-and-project} with Gaussian noise having standard deviation $\sigma=\frac{\sqrt{4\log(2/\delta)}}{\eps\cdot \card{D}}$ and projection set
    \begin{align*}
        \cS'_{2k,n}&:=\\
        &\Set{X\in (\R^n)^{\tensor k}\suchthat \exists \mu \in \Omega_{2k,n} \textnormal{ s.t. }\tilde{\E}_\mu \Brac{\tensorpower{x}{k}}=X}
    \end{align*}
    with $T^*$ as input.
    Let $\mathbf{T}$ be the algorithm's output.
    By \cref{fact:sos-injective-norm-certificate}, $\cG(\cS'_{2k,n})$ is bounded by $n^{k/4}$ if $k$ is even and by $\Paren{n^k\cdot k\cdot \log n}^{1/4}$ if $k$ is odd.
    Hence \cref{theorem:guarantees-add-noise-and-project} yields for $k$ even
    \begin{align*}
        \E \Snormt{\mathbf{T}-T^*}\leq O\Paren{ \frac{\sqrt{\log(2/\delta)}}{\eps\cdot \card{D}}\cdot n^{k/4}}\,,
    \end{align*}
    for $k$ odd
    \begin{align*}
    \E \Snormt{\mathbf{T}-T^*}\leq O\Paren{\frac{\sqrt{\log(2/\delta)}}{\eps\cdot \card{D}}\cdot\Paren{n^{k}\cdot k\cdot \log n}^{1/4}}\,.
    \end{align*}
    Rescaling, the dense part of the Theorem follows.
    
    For the sparse settings, let $T^*:= \frac{1}{\card{D}}\sum_{e\in \cE}m_{D}(e)\tensorpower{(e/\sqrt{t})}{k} \,.$
    Again this ensures the $\ell^2_2$ sensitivity is $4/\card{D}^2$ and $\Snormt{T^*}=1$
   Let $\cS_{4k-4,n}$ be the set defined in \cref{theorem:sos-sparse-certificates} (i.e. not related to the set defined above in the proof) .
    Notice $T^*$ yields a valid distribution over solutions to \cref{equation:sparse-tpca-program} and hence $T^*\in \cS_{4k-4,n}\,.$
    We use \cref{algorithm:add-noise-and-project} with projection set $\cS_{4k-4,n}\,,$ standard deviation $\frac{\sqrt{4\log(2/\delta)}}{\eps\cdot \card{D}}$ and the rescaled tensor $T^*$ as input.
    Combining \cref{theorem:guarantees-add-noise-and-project} and \cref{theorem:sos-sparse-certificates} we get for $k$ even
    $$\Snormt{\mathbf{T}-T^*}\leq O\Paren{ \frac{\sqrt{\log(2/\delta)}}{\eps\cdot \card{D}}\cdot t^{k/2}\cdot \sqrt{k\log\frac{n}{t^2}}}\,,$$
    for $k$ odd
    $$\Snormt{\mathbf{T}-T^*}\leq O\Paren{\frac{\sqrt{\log(2/\delta)}}{\eps\cdot \card{D}}\cdot t^{k/2}\cdot \sqrt{k\log n}}\,.$$
    Rescaling the result follows.
\end{proof}

%% file: background.tex
\section{Background}\label{section:background}

We introduce here background notions required in other sections of the paper.

The Gaussian complexity of a set is defined as follows:
\begin{definition}[Gaussian complexity]\label{definition:gaussian-complexity}
    Let $\cS\subseteq\R^n$ be a set, the Gaussian complexity of $\cS$ is defined as
\begin{align*}
    \cG(\cS):=\E_{\mathbf{W}\sim N(0,\Id_n)}\max_{v\in \cS}\iprod{\mathbf{W}, v}\,.
\end{align*}
\end{definition}

\subsection{Differential privacy}\label{section:preliminaries-privacy}

In this section we introduce standard notions of differential privacy \cite{dwork2006calibrating}. Unless specified otherwise, we use the notion of adjacency defined below.

\begin{definition}[Adjacent vectors]\label{definition:adjacent-vectors}
   Two vectors $u,v\in \R^n$ are said to be \textit{adjacent} if
    \begin{align*}
        \Normt{v-u}\leq 1\,.
    \end{align*}
\end{definition}

\begin{definition}[Differential privacy]\label{definition:differential-privacy}
	An algorithm $\cM:\cV\rightarrow\cO$ is said to be $(\eps, \delta)$-differentially private for $\eps, \delta >0$ if and only if, for every $S\subseteq \cO$ and every neighboring $u,v \in \cV$ we have 
	\begin{align*}
		\bbP \Brac{\cM(u)\in S}\leq e^\eps\cdot \bbP \Brac{\cM(v)\in S}+\delta\,.
	\end{align*}
\end{definition}

Differential privacy is closed under post-processing and composition.

\begin{lemma}[Post-processing]\label{lemma:differential-privacy-post-processing}
	If $\cM:\cV\rightarrow \cO$ is an $(\eps, \delta)$-differentially private algorithm and $\cM':\cO\rightarrow \cO'$ is any randomized function. Then the algorithm $\cM'\Paren{\cM(v)}$ is $(\eps, \delta)$-differentially private.
\end{lemma}


\subsubsection{Basic differential privacy mechanisms}\label{section:gaussian-mechanisms}

The Gaussian mechanism is among the most widely used mechanisms in differential privacy. It works by adding a noise drawn from the Gaussian distribution to the output of the function one wants to privatize. The magnitude of the noise depends on the sensitivity of the function.

\begin{definition}[Sensitivity of function]\label{definition:sensitivity}
	Let $f:\cY\rightarrow \R^n$ be a function, its $\ell_2$-sensitivity is 
	\begin{align*}
		\Delta_{f, 2}:= \max_{\substack{Y\,, Y' \in \cY\\ Y\,, Y'\text{ are adjacent}}}\Normt{f(Y)-f(Y')}\,.
	\end{align*}
\end{definition}

The Gaussian mechanism provides privacy proportional to the $\ell_2$-sensitivity of the function.

\begin{lemma}[Gaussian mechanism]\label{lemma:gaussian-mechanism}
		Let $f:\cY\rightarrow \R^n$ be any function with $\ell_2$-sensitivity at most $\Delta_{f,2}$. Let $0< \eps\,, \delta\leq 1$. Then the algorithm that adds $N\Paren{0, \frac{\Delta_{f, 2}^2 \cdot 2\log(2/\delta)}{\eps^2}\cdot \Id_n}$  to $f$ is $(\eps, \delta)$-DP.
\end{lemma}

\subsection{Sum-of-squares and pseudo-distributions}\label{section:preliminaries-sos-pseudodistributions}
We present here necessary background notion about the sum-of-squares framework. We borrow the description and all statements in this section from~\cite{d2020sparse, d2023higher}.

Let $w = (w_1, w_2, \ldots, w_n)$ be a tuple of $n$ indeterminates and let $\R[w]$ be the set of polynomials with real coefficients and indeterminates $w,\ldots,w_n$.
We say that a polynomial $p\in \R[w]$ is a \emph{sum-of-squares (sos)} if there are polynomials $q_1,\ldots,q_r$ such that $p=q_1^2 + \cdots + q_r^2$.

\subsubsection{Pseudo-distributions}\label{section:preliminaries-pseudodistributions}
Pseudo-distributions are generalizations of probability distributions.
We can represent a discrete (i.e., finitely supported) probability distribution over $\R^n$ by its probability mass function $\mu\from \R^n \to \R$ such that $\mu \geq 0$ and $\sum_{w \in \mathrm{supp}(\mu)} \mu(w) = 1$.
Similarly, we can describe a pseudo-distribution by its mass function.
Here, we relax the constraint $\mu\ge 0$ and only require that $\mu$ passes certain low-degree non-negativity tests.

Concretely, a \emph{level-$\ell$ pseudo-distribution} is a finitely-supported function $\mu:\R^n \rightarrow \R$ such that $\sum_{w} \mu(w) = 1$ and $\sum_{w} \mu(w) f(w)^2 \geq 0$ for every polynomial $f$ of degree at most $\ell/2$.
(Here, the summations are over the support of $\mu$.)
A straightforward polynomial-interpolation argument shows that every level-$\infty$-pseudo distribution satisfies $\mu\ge 0$ and is thus an actual probability distribution.
We define the \emph{pseudo-expectation} of a function $f$ on $\R^n$ with respect to a pseudo-distribution $D$, denoted $\tilde{\E}_{\mu(w)} f(w)$, as
\begin{equation}
	\tilde{\E}_{\mu(w)} f(w) = \sum_{w} \mu(w) f(w) \,\mper
\end{equation}
The degree-$\ell$ moment tensor of a pseudo-distribution $D$ is the tensor $\E_{\mu(w)} (1,w_1, w_2,\ldots, w_n)^{\otimes \ell}$.
In particular, the moment tensor has an entry corresponding to the pseudo-expectation of all monomials of degree at most $\ell$ in $w$.
The set of all degree-$\ell$ moment tensors of probability distribution is a convex set.
Similarly, the set of all degree-$\ell$ moment tensors of degree $d$ pseudo-distributions is also convex.
Key to the algorithmic utility of pseudo-distributions is the fact that while there can be no efficient separation oracle for the convex set of all degree-$\ell$ moment tensors of an actual probability distribution, there's a separation oracle running in time $n^{O(\ell)}$ for the convex set of the degree-$\ell$ moment tensors of all level-$\ell$ pseudodistributions.

\begin{fact}[\cite{shor1987quadratic,parrilo2000structured,nesterov2000squared,lasserre2001new}]
	\label[fact]{fact:sos-separation-efficient}
	For any $n,\ell \in \N$, the following set has a $n^{O(\ell)}$-time weak separation oracle (in the sense of \cite{grotschel1981ellipsoid}):
	\begin{equation}
		\Set{ \tilde{\E}_{\mu(w)} (1,w_1, w_2, \ldots, w_n)^{\otimes \ell} \mid \text{ degree-$\ell$ pseudo-distribution $\mu$ over $\R^n$}}\,\mper
	\end{equation}
\end{fact}
This fact, together with the equivalence of weak separation and optimization \cite{grotschel1981ellipsoid} allows us to efficiently optimize over pseudo-distributions (approximately)---this algorithm is referred to as the sum-of-squares algorithm.

The \emph{level-$\ell$ sum-of-squares algorithm} optimizes over the space of all level-$\ell$ pseudo-distributions that satisfy a given set of polynomial constraints---we formally define this next.

\begin{definition}[Constrained pseudo-distributions]
	Let $\mu$ be a level-$\ell$ pseudo-distribution over $\R^n$.
	Let $\cA = \{f_1\ge 0, f_2\ge 0, \ldots, f_m\ge 0\}$ be a system of $m$ polynomial inequality constraints.
	We say that \emph{$\mu$ satisfies the system of constraints $\cA$ at degree $r$}, denoted $\mu \sdtstile{r}{} \cA$, if for every $S\subseteq[m]$ and every sum-of-squares polynomial $h$ with $\deg h + \sum_{i\in S} \max\{\deg f_i,r\}\leq \ell$,
	\begin{displaymath}
		\tilde{\E}_{\mu} h \cdot \prod _{i\in S}f_i  \ge 0\,.
	\end{displaymath}
	We write $\mu\sdtstile{}{} \cA$ (without specifying the degree) if $\mu \sdtstile{0}{} \cA$ holds.
	Furthermore, we say that $\mu\sdtstile{r}{}\cA$ holds \emph{approximately} if the above inequalities are satisfied up to an error of $2^{-n^\ell}\cdot \norm{h}\cdot\prod_{i\in S}\norm{f_i}$, where $\norm{\cdot}$ denotes the Euclidean norm\footnote{The choice of norm is not important here because the factor $2^{-n^\ell}$ swamps the effects of choosing another norm.} of the coefficients of a polynomial in the monomial basis.
\end{definition}

We remark that if $\mu$ is an actual (discrete) probability distribution, then we have  $\mu\sdtstile{}{}\cA$ if and only if $\mu$ is supported on solutions to the constraints $\cA$.

We say that a system $\cA$ of polynomial constraints is \emph{explicitly bounded} if it contains a constraint of the form $\{ \|w\|^2 \leq M\}$.
The following fact is a consequence of \cref{fact:sos-separation-efficient} and \cite{grotschel1981ellipsoid},

\begin{fact}[Efficient Optimization over Pseudo-distributions]\label{fact:running-time-sos}
	There exists an $(n+ m)^{O(\ell)} $-time algorithm that, given any explicitly bounded and satisfiable system\footnote{Here, we assume that the bit complexity of the constraints in $\cA$ is $(n+m)^{O(1)}$.} $\cA$ of $m$ polynomial constraints in $n$ variables, outputs a level-$\ell$ pseudo-distribution that satisfies $\cA$ approximately. 
\end{fact}

\subsubsection{Sum-of-squares proof}\label{section:preliminaries-sos}
Let $f_1, f_2, \ldots, f_r$ and $g$ be multivariate polynomials in $w$.
A \emph{sum-of-squares proof} that the constraints $\{f_1 \geq 0, \ldots, f_m \geq 0\}$ imply the constraint $\{g \geq 0\}$ consists of  sum-of-squares polynomials $(p_S)_{S \subseteq [m]}$ such that
\begin{equation}
	g = \sum_{S \subseteq [m]} p_S \cdot \Pi_{i \in S} f_i
	\mper
\end{equation}
We say that this proof has \emph{degree $\ell$} if for every set $S \subseteq [m]$, the polynomial $p_S \Pi_{i \in S} f_i$ has degree at most $\ell$.
If there is a degree $\ell$ SoS proof that $\{f_i \geq 0 \mid i \leq r\}$ implies $\{g \geq 0\}$, we write:
\begin{equation}
	\{f_i \geq 0 \mid i \leq r\} \sststile{\ell}{}\{g \geq 0\}
	\mper
\end{equation}

Sum-of-squares proofs satisfy the following inference rules.
For all polynomials $f,g\colon\R^n \to \R$ and for all functions $F\colon \R^n \to \R^m$, $G\colon \R^n \to \R^k$, $H\colon \R^{p} \to \R^n$ such that each of the coordinates of the outputs are polynomials of the inputs, we have:

\begin{align}
	&\frac{\cA \sststile{\ell}{} \{f \geq 0, g \geq 0 \} } {\cA \sststile{\ell}{} \{f + g \geq 0\}}, \frac{\cA \sststile{\ell}{} \{f \geq 0\}, \cA \sststile{\ell'}{} \{g \geq 0\}} {\cA \sststile{\ell+\ell'}{} \{f \cdot g \geq 0\}} \tag{addition and multiplication}\\
	&\frac{\cA \sststile{\ell}{} \cB, \cB \sststile{\ell'}{} C}{\cA \sststile{\ell \cdot \ell'}{} C}  \tag{transitivity}\\
	&\frac{\{F \geq 0\} \sststile{\ell}{} \{G \geq 0\}}{\{F(H) \geq 0\} \sststile{\ell \cdot \deg(H)} {} \{G(H) \geq 0\}} \tag{substitution}\mper
\end{align}

Low-degree sum-of-squares proofs are sound and complete if we take low-level pseudo-distributions as models.

Concretely, sum-of-squares proofs allow us to deduce properties of pseudo-distributions that satisfy some constraints.

\begin{fact}[Soundness]
	\label{fact:sos-soundness}
	If $\mu \sdtstile{r}{} \cA$ for a level-$\ell$ pseudo-distribution $\mu$ and there exists a sum-of-squares proof $\cA \sststile{r'}{} \cB$, then $\mu \sdtstile{r\cdot r'+r'}{} \cB$.
\end{fact}

If the pseudo-distribution $\mu$ satisfies $\cA$ only approximately, soundness continues to hold if we require an upper bound on the bit-complexity of the sum-of-squares $\cA \sststile{r'}{} B$  (number of bits required to write down the proof).

In our applications, the bit complexity of all sum of squares proofs will be $n^{O(\ell)}$ (assuming that all numbers in the input have bit complexity $n^{O(1)}$).
This bound suffices in order to argue about pseudo-distributions that satisfy polynomial constraints approximately.

The following fact shows that every property of low-level pseudo-distributions can be derived by low-degree sum-of-squares proofs.

\begin{fact}[Completeness]
	\label{fact:sos-completeness}
	Suppose $d \geq r' \geq r$ and $\cA$ is a collection of polynomial constraints with degree at most $r$, and $\cA \vdash \{ \sum_{i = 1}^n w_i^2 \leq B\}$ for some finite $B$.
	
	Let $\{g \geq 0 \}$ be a polynomial constraint.
	If every degree-$d$ pseudo-distribution that satisfies $\mu \sdtstile{r}{} \cA$ also satisfies $\mu\sdtstile{r'}{} \{g \geq 0 \}$, then for every $\epsilon > 0$, there is a sum-of-squares proof $\cA \sststile{d}{} \{g \geq - \epsilon \}$.
\end{fact}

\subsubsection{Sum-of-squares toolkit}\label{section:sos-toolkit}

Here we present some well-known sum-of-squares bound that we will use throughout the paper.

\begin{fact}[Cauchy-Schwarz for pseudo-distributions \cite{barak2012hypercontractivity}]\label{fact:pseudo-cauchy-schwarz}
	Let $f,g$ be vector polynomials of degree at most $\ell$ in indeterminate $x\in \R^n$. Then, for any degree $2\ell$ pseudo-distribution $\mu$,
	\begin{align*}
		\tilde{\E}_\mu\Brac{\iprod{f,g}}\leq \sqrt{\tilde{\E}_\mu\brac{\norm{f}^2} }\cdot \sqrt{\tilde{\E}_\mu\brac{\norm{g}^2} }\,.
	\end{align*}
\end{fact}

We will also repeatedly use the following SoS version of Cauchy-Schwarz inequality and its generalization, Hölder's inequality:
\begin{fact}[Sum-of-Squares Cauchy-Schwarz]
	Let $x,y \in \R^d$ be indeterminites. Then,
	\[ 
	\sststile{4}{x,y} \Set{\Paren{\sum_i x_i y_i}^2 \leq \Paren{\sum_i x_i^2} \Paren{\sum_i y_i^2}} 
	\]
	\label{fact:sos-cauchy-schwarz}
\end{fact} 

We will use the following fact that shows how spectral certificates are captured within the SoS proof system.

\begin{fact}[Spectral Certificates] \label{fact:sos-spectral-certificates}
	For any $m \times m$ matrix $A$, 
	\[
	\sststile{2}{u} \Set{ \iprod{u,Au} \leq \Norm{A} \Norm{u}_2^2}\mper
	\]
\end{fact}

For sparse vectors, the statement below is often tighter.

\begin{fact}[Spare vector certificates \cite{d2020sparse}]\label{fact:sos-sparse-certificate}
    Let $\bm W\sim N(0,1)^{\times n}$ and let $$\cA:=\Set{X\in \R^{n\times n}\suchthat X\sge 0\,, \Tr X=1\,, \Normo{X}\leq t}\,.$$ Then with high probability over $\mathbf{W}$
    \begin{align*}
        \max_{X\in \cX}\iprod{X, \mathbf{W}}\leq O\Paren{t\sqrt{\log n/t^2}}\,.
    \end{align*}
\end{fact}

The next statement captures known certificates for the injective norms of random tensors.

\begin{fact}[\cite{hopkins2015tensor, d2023higher}]\label{fact:sos-injective-norm-certificate}
	Let $p\ge 3$ be an odd number, and let $\bm W \in \Paren{\R^{n}}^{\otimes p}$ be a tensor with i.i.d. entries from $N(0,1)$. 
	Then with probability $1-\delta$ (over $\bm W$) every pseudo-distribution $\mu$ of degree at least $2p-2$ on indeterminates $x=(x_1,\ldots,x_n)$ satisfies
	\[
	\tilde{\E}\iprod{x^{\otimes p}, \bm W} \le C\cdot \Paren{\Paren{n^p \cdot p \cdot \ln n}^{1/4} + 
		n^{p/4} \Paren{\ln(1/\delta)}^{1/4} + n^{1/4}\Paren{\ln(1/\delta)}^{3/4}}
	\cdot \Paren{\tilde{\E}\norm{x}^{2p-2}}^{\frac{p}{2p-2}}
	\] 
	for some absolute constant $C$.
\end{fact}